\documentclass[journal]{IEEEtran}
\usepackage{amsmath,amsfonts}
\usepackage{algorithmic}
\usepackage{algorithm}
\usepackage{array}
\usepackage[caption=false,font=normalsize,labelfont=sf,textfont=sf]{subfig}
\usepackage{textcomp}
\usepackage{stfloats}
\usepackage{url}
\usepackage{verbatim}
\usepackage{graphicx}
\usepackage{float}
\usepackage{cite}
\usepackage{tabularx}
\usepackage{threeparttable}
\usepackage{booktabs}
\usepackage{multirow}
\usepackage[hidelinks]{hyperref}
\usepackage{orcidlink}
\DeclareMathOperator{\cnorm}{CNorm}
\newcommand{\norm}[1]{\lVert #1\rVert}

\begin{document}

\title{Teaching a Robot Dog New Tricks: Diverse Quadruped Skills via Combined Reinforcement and Imitation Learning with Adversarial Task Selection}

\author{Lemon Foxmere\orcidlink{0009-0001-4299-3843}, Anthony Furman\orcidlink{0009-0009-9439-5325}, Yizheng Du\orcidlink{0009-0006-4384-503X}, Oliver Chang\orcidlink{0009-0000-7768-9887}, Leilani Gilpin\orcidlink{0000-0002-9741-2014}, and Steve McGuire\orcidlink{0000-0003-4650-7950}
\thanks{This work has been submitted to the IEEE for possible publication. Copyright may be transferred without notice, after which this version may no longer be accessible.}%
\thanks{\textit{(Lemon Foxmere and Anthony Furman are co-first authors.) }}%
\thanks{\textit{(Corresponding authors: Lemon Foxmere; Anthony Furman.)}}%
\thanks{Lemon Foxmere, Anthony Furman, Yizheng Du, and Steve McGuire are with the HARE Lab, University of California at Santa Cruz, Santa Cruz, CA 95064 USA (e-mail: lemonfoxmere@ucsc.edu; apfurman@ucsc.edu; ydu66@ucsc.edu; stmcguir@ucsc.edu).}%
\thanks{Oliver Chang and Leilani Gilpin are with the AIEA Lab, University of California at Santa Cruz, Santa Cruz, CA 95064 USA (e-mail: elochang@ucsc.edu; lgilpin@ucsc.edu).}%
}

\maketitle

\begin{abstract}
Reinforcement Learning (RL) has enabled legged robots to perform a range of skills in single-task settings. However, applications such as farm robotics or space exploration require diverse skills such as locomotion, digging, or close-range surveying. Training an end-to-end policy to address this problem remains difficult due to challenges such as sample inefficiency and gradient conflict between tasks in multi-task learning. We propose a three-stage method that trains a single policy to perform distinct tasks such as walking, digging, and hopping, and compose them into novel behaviors such as crawling. First, multiple teacher policies are trained using RL on narrowly defined tasks. Then, two additional stages train a student policy with a multi-teacher distillation setup that uses a combined RL and Imitation Learning (IL) objective under an adversarial task selection process that focuses training on the worst-performing task. With this method, we train a student policy that performs 22 tasks using 8 teachers. Evaluations show our method preserves motion quality and tracks commands more accurately than PPO and distill-then-finetune baselines, and in some cases generalizes to new tasks without explicit training. Finally, we demonstrate real-world robustness by deploying the resulting policy on a Unitree B1 quadruped.

\noindent Video: \textit{\href{https://youtu.be/V9yX04EBcFA}{https://youtu.be/V9yX04EBcFA}}
\end{abstract}

\begin{IEEEkeywords}
Hybrid Learning, Multi-Task Reinforcement Learning, Legged Robotics, Reinforcement Learning, Sim-to-Real Transfer
\end{IEEEkeywords}

\begin{figure*}[t]
\centering
\includegraphics[width=\textwidth]{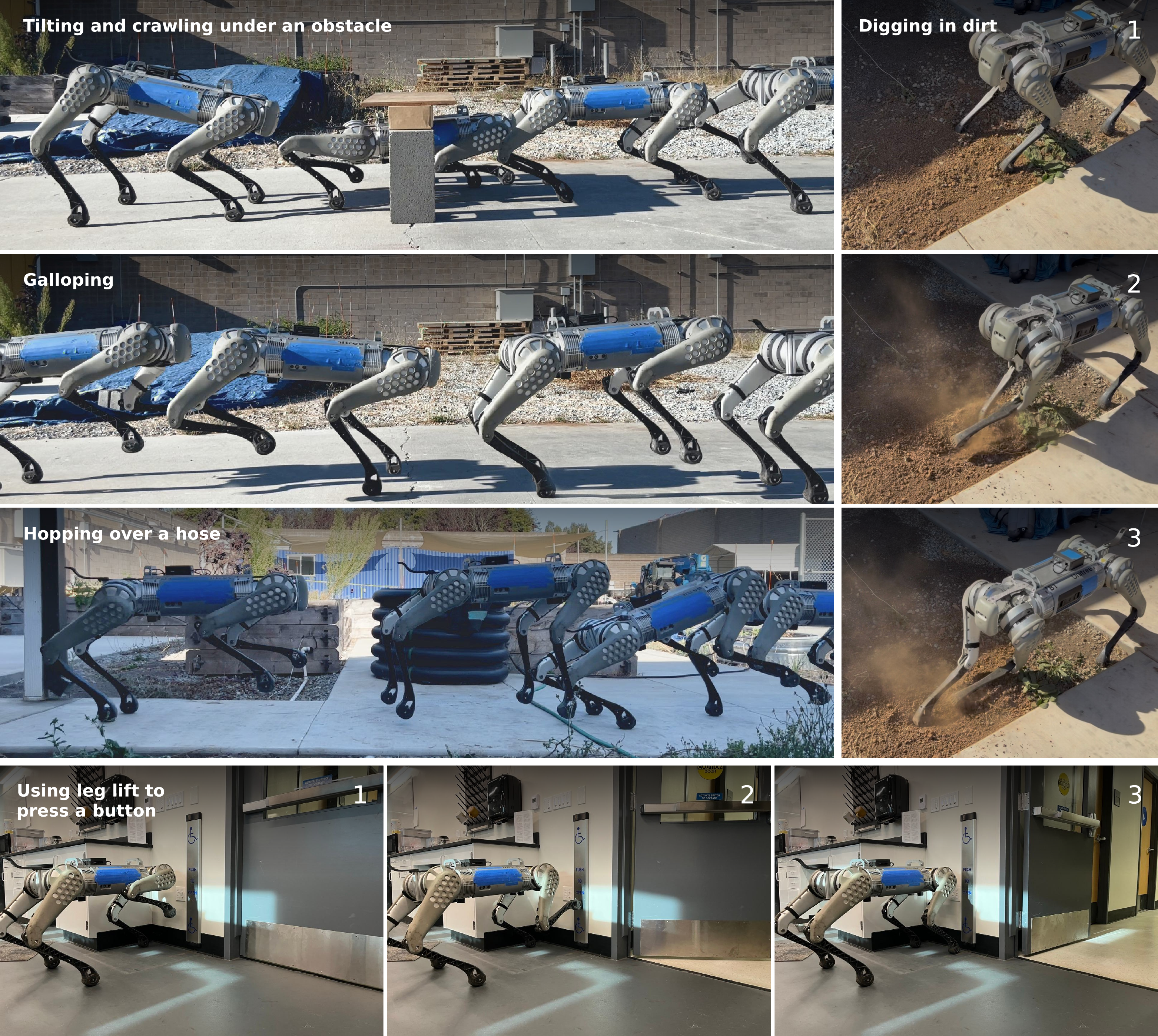}
\caption{Deployment of our policy on a Unitree B1 quadruped in real life. The robot is capable of performing a variety of tasks, including crawling under obstacles, galloping, hopping, digging, tilting, and lifting its leg to press buttons. A single policy is used in all deployments.}
\label{fig:irl_robots}
\vspace{-1em}
\end{figure*}
\section{Introduction}
\IEEEPARstart{R}{einforcement} Learning (RL) has gained much popularity in the domain of legged robotics in recent years, enabling robots to perform a diverse range of challenging tasks in single-task, real-world settings \cite{caluwaerts2023barkourbenchmarkinganimallevelagility, Miki_2022, cheng2023extremeparkourleggedrobots}. Despite this, the use of such robots remains limited in practice; applications like farm robotics and space exploration require many diverse skills such as locomotion, digging, and close-range surveying. One approach to this problem, known as Multi-task Reinforcement Learning (MTRL), is to train a single end-to-end policy that can perform many distinct tasks on command. However, doing so remains an open challenge. RL's sample inefficiency means that many techniques are often limited to learning only one task per policy, or tasks that have substantial shared structure \cite{xing2024multitaskreinforcementlearningquadrotors, naveen2026learningadaptrepresentationbasedreinforcement}. Reward gaming makes training difficult and tedious, and can yield policies that do not produce desirable behaviors \cite{amodei2016concreteproblemsaisafety}. Finally, challenges such as gradient conflict and catastrophic forgetting hinder effective policy convergence \cite{yu2020gradientsurgerymultitasklearning, varghese2020survey}.

Imitation learning (IL) offers an alternative approach, circumventing several of RL's core difficulties \cite{hussein2017imitation, mirza2025imitation, rajeswaran2018dapg}. Reference motions from motion capture data have been successfully applied to create realistic legged locomotion \cite{escontrela2022amp}, and many works use synthetic demonstrations produced by ``teacher'' policies trained with RL to distill a student model \cite{rusu2016policydistillation, zhao2026adaptivehumanoidcontrolmultibehavior, rudin2025parkourwildlearninggeneral}.

Given the strengths and limitations of each method, combining the two paradigms has been studied extensively \cite{lu2023imitationenoughrobustifyingimitation, rajeswaran2018dapg, rudin2025parkourwildlearninggeneral}, with IL providing training stability and behavioral fidelity and RL enabling robust generalization to out-of-distribution (OOD) states \cite{lu2023imitationenoughrobustifyingimitation, rudin2025parkourwildlearninggeneral}. Notably, Rudin et al.~\cite{rudin2025parkourwildlearninggeneral} adopt a Distill-then-Finetune (DtF) structure, first pre-training a strong student policy through multi-teacher distillation, then finetuning it with RL to learn new tasks and improve on existing ones. The resulting policy demonstrates robust performance in rough terrain locomotion.

\subsection{Contribution}

Building on this DtF structure, we propose a three-stage training method for effectively learning multiple distinct tasks, performing them on command, and composing multiple tasks to form novel behaviors. Under our method, we first train multiple teacher policies to perform narrowly defined tasks using RL. These teachers are used as the imitation target in our optimization objective, and are trained to prioritize natural movements over perfect task performance. Then, in two additional stages of training, we optimize a student policy over a combined IL and RL objective. Finally, to allocate training effort across tasks of uneven difficulty, we introduce an adversarial task selection strategy that preferentially samples commands on which the student currently performs worst. 

The resulting policy is capable of learning multiple distinct tasks from its teachers and composing several tasks together to create new behaviors. For example, a policy trained for body height tracking and omni-directional walking can learn to compose the two to create a crawling motion (i.e., walking at a low height). We demonstrate this capability by training a student model to perform 22 total tasks: 8 taught by the teacher policies and 14 learned by composing two teacher tasks at once. Evaluations show that policies trained under this method preserve motion quality and track commands more accurately than PPO and DtF baselines, and exhibit minimal effects of reward gaming. The policies also generalize well to tasks beyond what is trained, meaningfully composing three or more tasks despite never encountering such compositions during training. Finally, we show that these policies exhibit robustness in real-world deployment while maintaining high motion quality and command responsiveness.

\section{Related Work}

Model distillation, first introduced to compress model sizes, was later used to distill domain-specific knowledge from expert models to a single model \cite{bucila2006modelcompression, hinton2015distilling}. Following Rusu et al.~\cite{rusu2016policydistillation}, model distillation was popularized in the context of RL, where one or more expert models trained using RL can distill their knowledge into a single student model. However, the student often suffers from degraded performance compared to any given teacher \cite{rusu2016policydistillation}. Using multi-teacher distillation, Caluwaerts et al.~\cite{caluwaerts2023barkourbenchmarkinganimallevelagility} distilled several specialist quadruped locomotion policies into a single generalist policy and demonstrated success on their Barkour benchmark.

A parallel line of work addresses this performance gap by combining IL and RL into a joint objective. Rajeswaran et al.~\cite{rajeswaran2018dapg} introduced DAPG, which augments a standard RL policy gradient objective with a behavioral cloning (BC) auxiliary loss, allowing demonstrations to guide early exploration while RL drives final performance as the demonstration term decays. Lu et al.~\cite{lu2023imitationenoughrobustifyingimitation} extend this formulation to autonomous driving at scale by using a combined BC and soft actor-critic objective (BC-SAC), demonstrating that the BC auxiliary loss preserves naturalistic behavior while the RL objective provides robustness in OOD scenarios. Similarly, Schmitt et al.~\cite{schmitt2018kickstarting} add a multi-teacher distillation loss to the RL objective but reduce its weight over training. In contrast, our method retains a nonzero imitation weight throughout training so that the BC term continues to anchor motion quality. Another approach to address motion quality is Adversarial Motion Priors (AMP), which replaces style-rewards with a discriminator trained on reference motion~\cite{peng2021amp}. AMP has since been successfully demonstrated in quadruped locomotion, producing natural and energy-efficient motion~\cite{escontrela2022amp}. DAgger-style online data collection \cite{ross2011reductionimitationlearningstructured}, where teachers are queried under the student's own state distribution rather than a fixed offline dataset, further addresses distributional shift during online learning by keeping demonstration data on-policy.

In the context of MTRL, problems such as gradient conflicts \cite{yu2020gradientsurgerymultitasklearning}, sample inefficiency, and catastrophic forgetting all remain obstacles that hinder effective policy optimization \cite{varghese2020survey}. Methods such as PCGrad address gradient conflicts in MTRL by minimizing contradictory gradient signals through gradient projections, and have seen success in MTRL training on humanoid robots \cite{yu2020gradientsurgerymultitasklearning, zhao2026adaptivehumanoidcontrolmultibehavior}. Techniques combining multi-teacher distillation with RL have since been demonstrated on quadrupedal platforms as an effective solution for tasks such as rough terrain locomotion \cite{rudin2025parkourwildlearninggeneral, huang2025moelocomixtureexpertsmultitask}. Notably, in Rudin et al.~\cite{rudin2025parkourwildlearninggeneral}, task switching is context-driven and occurs automatically rather than through explicit command inputs, and task composition performance is highly dependent on training setups due to the student's tendency to overwrite previously acquired behaviors. Huang et al.~\cite{huang2025moelocomixtureexpertsmultitask} have demonstrated that a Mixture of Experts structure (MoELoco) can be used to realize multiple behaviors under one system, as well as compose novel behaviors from multiple experts. However, this method relies on a specialized gating network with multiple expert models, and is not an end-to-end solution.

\section{Methodology}

\subsection{Problem Definition}
We model the multi-task problem as a Markov Decision Process (MDP) defined by the tuple $(\mathcal{S}, \mathcal{A}, \mathcal{T}, \mathcal{R}, \gamma)$, where $\mathcal{S}$ is the state space, $\mathcal{A}$ is the action space, $\mathcal{T}$ is the transition function, $\mathcal{R}$ is the reward function, and $\gamma$ is the discount factor.

We define a ``single task'' to be a narrowly-defined behavior for which a dedicated teacher policy exists, and a ``composite task'' to be a task that requires the concurrent execution of two or more single tasks and for which no teacher policy exists. In this context, we define $\mathcal{X} = \{x_1, \dots, x_n\}$ to be a set of $n$ single tasks, and $\mathcal{Y} = \{y_1, \dots, y_m\}$ to be a set of $m$ single and composite tasks that the student policy will be trained to perform. Which tasks are combined is a decision based on the feasibility of the system dynamics and the requirements for robot performance.

To specify tasks and tracking commands for the student, we define a command vector $\mathbf{c}_\mathcal{Y}$ as a set of tracking targets that are shared across $\mathcal{Y}$. Tracking targets are scalar values that the robot must track, or binary flags for whether a task should be performed. Each task is associated with a group of targets that may overlap with other tasks. By construction, we can meaningfully command any task $y \in \mathcal{Y}$ by simply setting the necessary task targets.

Finally, the student policy $\pi_\theta(a \mid s, \mathbf{c}_\mathcal{Y})$ is conditioned on the environment state $s \in \mathcal{S}$ and the command vector $\mathbf{c}_\mathcal{Y}$, and is parameterized by $\theta$.

\begin{figure*}[t]
\centering
\includegraphics[width=\textwidth]{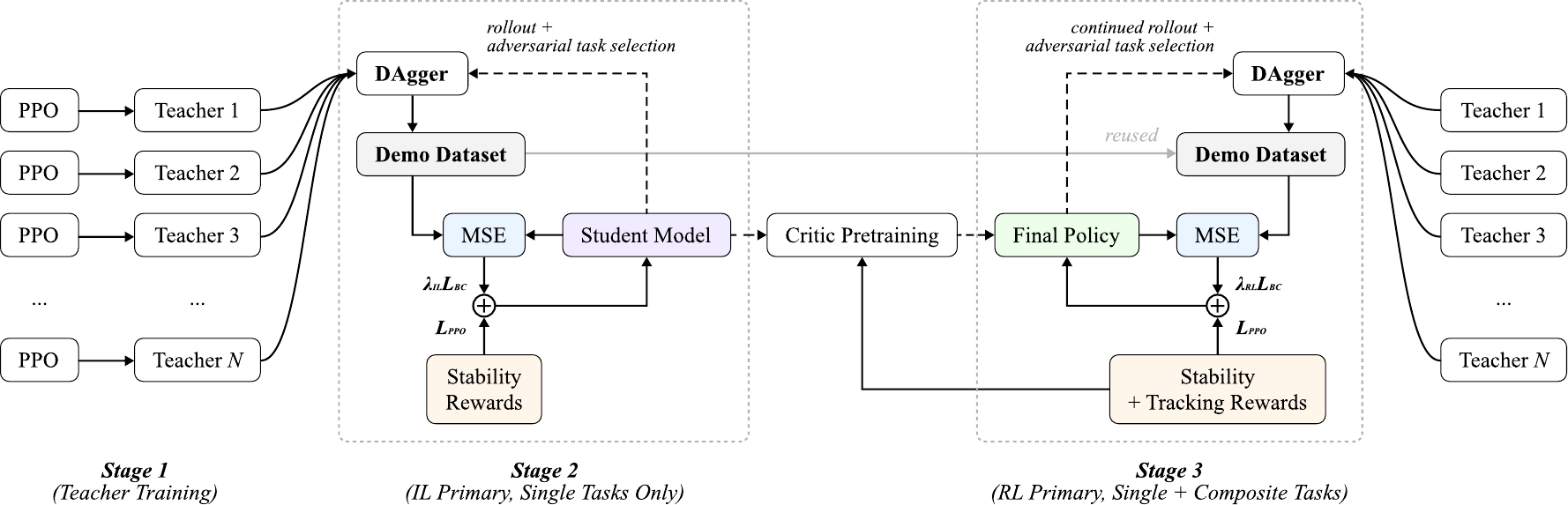}
\caption{An overview of our method. Teacher policies are first trained via PPO to perform each single task. During ILP, a student policy is optimized over a combined $L_\mathrm{PPO} - \lambda_\mathrm{IL} L_\mathrm{BC}$ objective, with the behavioral cloning demonstrations collected from the teachers via DAgger. In this stage, rewards only encourage body stability and motion smoothness. During RLP, a new critic is first pre-trained on the RLP reward landscape, and the student is optimized over the same combined objective but with a weaker $\lambda_\mathrm{RL}$. Here, rewards are extended to also finetune certain behaviors and encourage command tracking. Throughout both ILP and RLP, adversarial task selection is used during training to accumulate more transitions on the worst performing tasks.}
\label{fig:method_arch}
\vspace{-1em}
\end{figure*}
\subsection{Teacher Policies}
For each task $x_i \in \mathcal{X}$, a corresponding teacher policy $\pi_i^*(a \mid s, \mathbf{c}_i)$ is trained using proximal policy optimization (PPO) under a reward landscape specific to that task. Since teachers serve only to generate reference motion for the student policy to imitate, we focus teacher training on achieving the most natural-appearing motion rather than perfect command tracking performance. For this, teachers are trained on privileged information unavailable to the student, such as foot contact forces, yielding smooth and natural movements while still retaining good command tracking performance~\cite{chen2019learningcheating}. The final teacher policies are frozen prior to student training, and define the behavioral and tracking baseline that the student is expected to reach.

\subsection{Combined Training Objective}
The student is trained using a combined objective consisting of a PPO clipped surrogate objective and a BC loss.

The PPO clipped surrogate objective is defined as:
\begin{equation}
    \scalebox{0.9}{$L^{\text{PPO}}(\theta) = \mathbb{E}_t \left[ \min \left( r_t(\theta) \hat{A}_t,\ \text{clip}(r_t(\theta), 1-\epsilon, 1+\epsilon) \hat{A}_t \right) \right]$}
\end{equation}
where $r_t(\theta)$ is the probability ratio between the current and old policy, $\hat{A}_t$ is the estimated advantage at timestep $t$, and $\epsilon$ is the clipping threshold \cite{schulman2017proximalpolicyoptimizationalgorithms}.
 
The BC loss is defined as:
\begin{equation}
    L^{\text{BC}}(\theta) = \mathbb{E}_{(s, \mathbf{c}_\mathcal{Y}, a^*) \sim \mathcal{D}} \left[ \left\| \mu_\theta(s, \mathbf{c}_\mathcal{Y}) - a^* \right\|^2 \right]
\end{equation}
where $\mathcal{D}$ is the demonstration dataset collected via DAgger, $a^* = \pi_i^*(s, \mathbf{c}_i)$ is the action from the appropriate teacher policy for command $\mathbf{c}_\mathcal{Y}$, and $\mu_\theta$ is the mean of the student's action distribution \cite{ross2011reductionimitationlearningstructured}. If there exists no appropriate teacher policy to poll for a given $\mathbf{c}_\mathcal{Y}$, DAgger is skipped and no imitation data sample is collected for that point. MSE is used so that the imitation signal acts only on the policy mean and not the standard deviation (STD), leaving exploration under the control of PPO.
 
By combining both, we obtain the combined objective:
\begin{equation}\label{eq:objective}
    L(\theta) = L^{\text{PPO}}(\theta) - \lambda L^{\text{BC}}(\theta)
\end{equation}
where $\lambda > 0$ controls the relative weight of the imitation component. This will be our primary optimization target for the student policy.

\subsection{Imitation Primary and Reinforcement Primary}

When training the student policy, we split our training process into two stages: an imitation learning primary stage (ILP), and a reinforcement learning primary stage (RLP).

During ILP, we focus on training the student to obtain a baseline performance from the teachers. As such, we only train on single tasks and use a high $\lambda$ coefficient in $L(\theta)$ so that BC dominates PPO. RL rewards and gradients are still present, but only task-invariant stability terms are used, such as penalties for excessive joint power and rewards for body stability.

During RLP, we focus on improving previously learned single tasks and learning novel composite tasks. As such, we train on both single and composite tasks using a low $\lambda$ coefficient to allow PPO to drive learning. Here, the IL gradients are still present, but now act as an anchor to teacher policy behaviors so that motion quality is preserved as learning continues.

\subsection{Online Data Collection}
To keep demonstration data on-policy, we integrate a DAgger-style collection procedure \cite{ross2011reductionimitationlearningstructured} into the main PPO training loop. During both ILP and RLP, the student policy $\pi_\theta$ rolls out under a randomly sampled student command $\mathbf{c}_\mathcal{Y}$. At every timestep $t$, if the task commanded by $\mathbf{c}_\mathcal{Y}$ is a single task, the corresponding teacher policy $\pi_i^*$ is queried at the student's current state $s_t$ with $\mathbf{c}_{it}$, the equivalent command in its own format, to produce a demonstration action $a_t^* = \pi_i^*(s_t, \mathbf{c}_{it})$, and the tuple $(s_t, \mathbf{c}_\mathcal{Y}, a_t^*)$ is appended to $\mathcal{D}$. If no corresponding teacher exists, such as in cases of composite tasks, no samples are collected for that timestep $t$.
 
To prevent a complete collapse of the student's action distribution STD during ILP, zero-mean Gaussian noise is added to student actions throughout this stage~\cite{rudin2025parkourwildlearninggeneral}.

\subsection{Adversarial Task Selection}

While DAgger helps keep demonstration data on-policy, uniform task sampling in DAgger introduces a systematic bias under multi-task training. Tasks the student masters quickly survive longer before termination and thus contribute disproportionately more transitions to $\mathcal{D}$, while poorly performing tasks are undersampled relative to their difficulty.
 
To correct for this, we replace the uniform sampling with an adversarially weighted distribution that allocates more episodes to tasks with the largest performance deficit. We measure this deficit through a series of error channels $\mathcal{E} = \{e_1, \dots, e_l\}$, where each channel $e$ measures a component of the raw tracking error for some single task $x$. Each task $y_i$ therefore activates a subset $\mathcal{E}_i \subseteq \mathcal{E}$ containing the error channels relevant to measuring its performance. By construction, a composite task's active channels are the union of its constituents' active channels.

We then normalize all channels by defining a channel normalization function
\begin{equation}\label{eq:error_norm}
    \cnorm(e) = \tanh(a\frac{v_e}{\tau_e}), \quad a=\tanh^{-1}(0.2),
\end{equation}
where $\tau_e > 0$ and $v_e \ge 0$, $v_e$ is the error for channel $e$, and $\tau_e$ is the nominal error for channel $e$ taken as the error attained by its corresponding teacher model at the end of training. Doing so normalizes all error values to $[0,1)$ and has the added benefit of converting absolute errors into relative measurements of a model's deficit to baselines set by teacher policies. Thus, from this operation, we can obtain a deficit score $\bar E_i$ of task $y_i$ as:
\begin{equation}
    \bar{E}_i = \frac{1}{|\mathcal{E}_i|} \sum_{e \in \mathcal{E}_i} \cnorm(e)
\end{equation}
Finally, we obtain a task selection PMF over $\mathcal{Y}$ as
\begin{equation}
    p_i = \kappa \frac{\bar{E}_i^{\beta}}{\sum_{j=1}^{|\mathcal{Y}|} \bar{E}_j^{\,\beta}} + \frac{1 - \kappa}{|\mathcal{Y}|},
\end{equation}
where $\kappa \in [0,1]$ is a mixing coefficient interpolating between the adversarial component and the uniform prior $\tfrac{1}{|\mathcal{Y}|}$, and $\beta \ge 1$ is a sharpness exponent to prevent mid-range probability compression as $|\mathcal{Y}|$ increases.

\subsection{Critic Pretraining}

Once the student achieves stable performance in ILP, it transitions to RLP for finetuning and learning new composite tasks. However, directly starting updates with low $\lambda$ at this point causes training instability due to the critic being out of distribution with respect to the RLP environment. Thus, the critic becomes unreliable in value estimation and hinders policy convergence. Following Rudin et al.~\cite{rudin2025parkourwildlearninggeneral}, we address this issue with a critic pretraining stage prior to actor optimization.
 
During critic pretraining, the actor parameters are frozen and the policy STD is fixed at an initial value $\sigma_0$. We then train a new critic $V$ against the RLP rewards under the full task distribution. This stage is held until the critic reaches a stable level of performance, at which point the actor is unfrozen and RLP training fully begins.

\subsection{Reward Design}

During the ILP stage, we keep the reward landscape simple and limited to terms that encourage body stability and motion smoothness. This allows IL to drive most of the learning while RL encourages small deviations that improve motion smoothness and stability.

However, during the RLP stage, the reward landscape must be expanded to target minimizing tracking errors and encourage new task learning for task composition. Since the BC term already anchors the policy to teacher distributions, preserving behavioral quality, the reward landscape no longer needs to encode every aspect of desirable behaviors on its own, simplifying reward design significantly. In our scenario, while teacher training required around 18 to 34 reward terms per policy (26 on average) to achieve desirable motion behaviors and tracking performance, the RLP stage only required 32 reward terms total across all 22 tasks for effective policy convergence. Qualitatively, we found that tuning RLP rewards was less time-consuming than tuning the teacher rewards. 

In both landscapes, we keep reward term weights constant and do not adjust weighting based on curricula.

\begin{table}[t]
\centering

\captionsetup{justification=raggedright,singlelinecheck=false}
\caption{Symbols}
\label{tab:symbols}

\begin{tabularx}{\columnwidth}{@{}l >{\raggedright\arraybackslash}X@{}}
\toprule
\textbf{Symbol} & \textbf{Description}\\
\midrule
$\mathbf q, \mathbf q_\mathrm{lay}$ & Current joint positions and the robot ``off'' position.\\
$\mathbf v_b, \boldsymbol\omega_b$ & Body velocities in body frame.\\
$\mathbf v^*, \boldsymbol\omega^*$ & Commanded body velocities.\\
$\theta, \theta^*$ & World frame body pitch and the commanded pitch angle.\\
$\phi, \phi^*$ & Body yaw and commanded yaw, computed relative to the heading defined by hind foot positions.\\
$h, h^*$ & Stance height and commanded height, computed as the vertical distance between the body and the lowest foot.\\
$\mathbf{G}^*$ & A vector $\in \{0, 1\}^2$ that commands the gait to perform. $[0,0]$ for walk, $[1,0]$ for hop, and $[0,1]$ for gallop. $[1,1]$ is invalid.\\
$\mathbf{L}^*$ & A vector $\in \{0, 1\}^2$ that commands which of the two front legs is to be lifted. $[0,0]$ is no leg lift.\\
$F_f, F_z^*$ & Vertical contact force for foot $f \in \{\mathrm{FL}, \mathrm{FR}\}$ and the commanded dig force.\\
$\mathbf p^H_f, \mathbf p^{H*}_f$ & Foot $f$'s position and commanded position in frame of the hip it's attached to.\\
$\mathbf p^B_f$ & Foot $f$'s position in body frame.\\
\bottomrule
\end{tabularx}
\vspace{-1em}
\end{table}
\begin{table}
\centering
\captionsetup{justification=raggedright,singlelinecheck=false}
\caption{Single and Composite Task Descriptions with Commands}
\label{tab:task_and_desc}
\begin{tabularx}{\columnwidth}{@{}l X l@{}}
\toprule
Task & Behavioral Description & Commands \\
\midrule
\textbf{Walk} & Omni-directionally trots, alternating diagonal foot pairs. & $\mathbf v^*, \boldsymbol\omega^*$\\ 
\textbf{Gallop} & Omni-directionally bounds, alternating front then hind foot pairs. & $\mathbf v^*, \boldsymbol\omega^*$ \\
\textbf{Hop} & Omni-directionally hops, lifting and landing with all feet at once. & $\mathbf v^*, \boldsymbol\omega^*$ \\
\textbf{Body Height} & Crouches to a certain stance height. & $h^*$\\
\textbf{Body Tilt} & Pitches the body to a certain angle. & $\theta^*$ \\
\textbf{Body Yaw} & Twists body while keeping a rectangular stance. & $\phi^*$\\
\textbf{Leg Lift} & Lifts a leg to a specified position. & $\mathbf{L}^*, \mathbf p^{H*}_f$ \\
\textbf{Dig} & Scrapes the ground with both front feet in one area at a specified downward force. & $F_z^*$ \\
\midrule
\textbf{Walk+Tilt} & Walks with body tilted. & $\mathbf v^*, \boldsymbol\omega^*, \theta^*$  \\
\textbf{Walk+Height} & Walks while crouching. & $\mathbf v^*, \boldsymbol\omega^*, h^*$  \\
\textbf{Gallop+Tilt} & Gallops with body tilted. & $\mathbf v^*, \boldsymbol\omega^*,\theta^*$  \\
\textbf{Gallop+Height} & Gallops while crouching. & $\mathbf v^*, \boldsymbol\omega^*, h^*$ \\
\textbf{Hop+Tilt} & Hops with body tilted. & $\mathbf v^*, \boldsymbol\omega^*, \theta^*$ \\
\textbf{Hop+Height} & Hops while crouching. & $\mathbf v^*, \boldsymbol\omega^*, h^*$ \\
\textbf{Tilt+Height} & Tilts body while crouching.  & $\theta^*, h^*$ \\
\textbf{Tilt+Yaw} & Tilts and twists body. & $\theta^*, \phi^*$ \\
\textbf{Yaw+Height} & Twists body while crouching. & $\phi^*, h^*$ \\
\textbf{Lift+Tilt} & Lifts a leg with body tilted. & $\mathbf{L}^*, \mathbf p^{H*}_f, \theta^*$ \\
\textbf{Lift+Height} & Lifts a leg while crouching. & $\mathbf{L}^*, \mathbf p^{H*}_f, h^*$ \\
\textbf{Lift+Yaw} & Lifts a leg with body twisted. & $\mathbf{L}^*, \mathbf p^{H*}_f, \phi^*$ \\
\textbf{Dig+Tilt} & Digs with body tilted. & $F_z^*, \theta^*$ \\
\textbf{Dig+Yaw} & Digs with body twisted. & $F_z^*, \phi^*$ \\
\bottomrule
\end{tabularx}
\vspace{-1em}
\end{table}
\section{Experiments and Evaluations}

We evaluate the performance of our method on 8 single tasks and 14 non-trivial and dynamically feasible composite tasks. Each policy is trained on all 22 tasks, listed and described in Table~\ref{tab:task_and_desc}. For composite tasks, we only train on 2-task compositions. All compositions of three or more tasks are reserved for generalization evaluation. 

\subsection{Baseline Setup}

We compare our method against three baselines: Distill-then-Finetune (DtF) as used by Rudin et al.~\cite{rudin2025parkourwildlearninggeneral}, single-stage PPO, and a teacher router that selects the corresponding teacher policy for a given task.

For DtF, a student policy is first warm-started via multi-teacher DAgger distillation. A critic is then pre-trained, after which PPO finetunes the student on both single and composite tasks. DtF uses the same teacher policies, critic pretraining process, and student architecture as our method. Distillation and finetuning use the same number of timesteps as our ILP and RLP respectively. For PPO, we use the same model architecture and hyperparameters as our student where applicable, with the exception of an initial policy STD of 1.0 instead of 0.5 to match standard PPO formulations. For both DtF and PPO, we use the same reward landscape as in RLP. See Appendix Section~\ref{subsec:training_setup} for additional training details.

\subsection{Command Vector}

To command tasks and tracking targets, we define a 14-dimensional command vector $\mathbf{c}_\mathcal{Y}$ using notations and symbols defined in Table~\ref{tab:symbols}:
\begin{equation*}
    \mathbf{c}_\mathcal{Y} = [\mathbf{v}^*, \boldsymbol{\omega}^*, \theta^*, \phi^*, \mathbf{L}^*, \mathbf{p}^{H*}_f, \mathbf{G}^*, F^*_z, h^*]
\end{equation*}

A property of this layout is that all tasks and targets are commanded at all times, blurring the distinction  between single and composite tasks. For clarity, we set each task to vary only the command terms as defined in Table~\ref{tab:task_and_desc}. Terms that are not actively varied by the current task are set to default values as defined in Table~\ref{tab:cmd_defaults}. For example, to command walk, we set desired values for $\mathbf{v}^*$ and $\boldsymbol{\omega}^*$ in addition to default values for $h^*$, $\theta^*$, etc. Thus, the distinction between tasks is purely which command terms are actively varied. We also reiterate that a ``single task'' is any task for which a teacher was trained.

\begin{table}[t]
\centering

\captionsetup{justification=raggedright,singlelinecheck=false}
\caption{Command Default Values For Each Component}
\label{tab:cmd_defaults}

\begin{tabularx}{\columnwidth}{@{}l l >{\raggedright\arraybackslash}X@{}}
\toprule
\textbf{Components} & \textbf{Default} & \textbf{Notes}\\
\midrule
$\mathbf p^{H*}_f$ & $[0,0,0]$ &\\
$\mathbf v^*, \mathbf G^*, \mathbf L^*$ & $[0,0]$ & \\
$\theta^*, \boldsymbol \omega^*, \phi^*, F^*_z$ & $0$ & \\
$h^*$ & $0.55$ & Matches walk teacher training height\\
\bottomrule
\end{tabularx}
\vspace{-1em}
\end{table}
\begin{table}[t]
\centering

\captionsetup{justification=raggedright,singlelinecheck=false}
\caption{All Evaluation Channels}
\label{tab:channel_definitions}

\newcolumntype{L}{>{\raggedright\arraybackslash}X}
\begin{tabularx}{\columnwidth}{@{}l L l@{\hspace{1em}}}
\toprule
\textbf{Channel} & \textbf{Expression} & \textbf{Active Condition}\\
\midrule
\texttt{vx} & $|\mathbf{v}^*_x-\mathbf{v}_{b,x}|$ & Always\\
\texttt{vy} & $|\mathbf{v}^*_y-\mathbf{v}_{b,y}|$ & Always\\
\texttt{wz} & $|\boldsymbol{\omega}^*-\boldsymbol{\omega}_b|$ & Always\\
\texttt{height} & $| h^* - h |$ & $h^* > 0$\\
\texttt{sit\_pose} & $\norm{\mathbf{q}-\mathbf{q}_\mathrm{lay}}$ & $h^* = 0$\\
\texttt{tilt} & $|\theta^* - \theta|$ & Always\\
\texttt{foot\_pos} & $\norm{\mathbf{p}^{H*}_f - \mathbf{p}^H_f}$ & $\norm{\mathbf{L}^*} > 0$\\
\texttt{yaw} & $|\phi^* - \phi|$ & $\norm{[\mathbf{v}^*, \boldsymbol{\omega}^*]} = 0$\\
\texttt{dig\_force} & $\mathrm{mean}_{f\in\{\mathrm{FL}, \mathrm{FR}\}} |F^*_z - F_f|$ & $F^*_z > 0$\\
\texttt{dig\_dist} & $\mathrm{mean}_{f\in\{\mathrm{FL}, \mathrm{FR}\}} |\mathbf{p}^B_{f,y}|$ & $F^*_z > 0$\\
\bottomrule
\end{tabularx}
\vspace{-1em}
\end{table}
\begin{table}[t]
\centering

\caption{Explicit and Implicit Channels Per Single Task}
\label{tab:single_task_channels}

\newcolumntype{L}{>{\raggedright\arraybackslash}X}
\begin{tabularx}{\columnwidth}{@{}l L L@{}}
\toprule
\textbf{Task} & \textbf{Explicit Channels} & \textbf{Implicit Channels}\\
\midrule
Walk, Hop, Gallop & \texttt{vx}, \texttt{vy}, \texttt{wz} & \texttt{tilt}, \texttt{height}\\
\addlinespace
Body Height & \texttt{height}, \texttt{sit\_pose} & \texttt{vx}, \texttt{vy}, \texttt{wz}, \texttt{tilt}, \texttt{yaw}\\
\addlinespace
Body Tilt & \texttt{tilt} & \texttt{vx}, \texttt{vy}, \texttt{wz}, \texttt{height}, \texttt{yaw}\\
\addlinespace
Body Yaw & \texttt{yaw} & \texttt{vx}, \texttt{vy}, \texttt{wz}, \texttt{tilt}, \texttt{height}\\
\addlinespace
Leg Lift & \texttt{foot\_pos} & \texttt{vx}, \texttt{vy}, \texttt{wz}, \texttt{tilt}, \texttt{height}, \texttt{yaw}\\
\addlinespace
Dig & \texttt{dig\_force}, \texttt{dig\_dist} & \texttt{vx}, \texttt{vy}, \texttt{wz}, \texttt{tilt}, \texttt{height}, \texttt{yaw}\\
\bottomrule
\end{tabularx}
\vspace{-1em}
\end{table}
\subsection{Metrics and Scores}

When evaluating task performance, we record all evaluation channels for that task as defined in Table~\ref{tab:channel_definitions} and listed in Table~\ref{tab:single_task_channels} during each timestep. Each channel is derived from a significant RLP reward term used during training, and jointly the channels reflect the RLP reward landscape, its goals, and the desired outcome of each task.

Due to the nature of the command layout as described previously, these channels are split into two categories: explicit and implicit. More formally, a task's explicit channels are channels that are active and contain command terms the task explicitly commands, whereas its implicit channels are channels that are active and only contain command terms the task holds constant at the default value. Some channels, such as \texttt{dig\_dist} and \texttt{sit\_pose}, which do not contain command terms listed in Table~\ref{tab:task_and_desc}, are included as explicit channels in some tasks to measure behavioral quality and outcomes of the commanded terms. Finally, we also record policy observation and foot contact pattern per timestep.

From these channels, we compute an explicit and an implicit performance score for each task. For a given task, each channel's value is normalized using the $\cnorm$ function defined in (\ref{eq:error_norm}) with $e_n = 1 - \cnorm(e)$. Here, we set $a = \tanh^{-1}(0.5)$ and $\tau_e$ to the explicit channel values obtained by the corresponding teacher policy. Finally, the normalized values are averaged over explicit and implicit channels separately to produce an explicit and implicit score for each task. Each score is bounded to $[0,1]$, where 0.5 corresponds to teacher-level tracking and 1 to perfect tracking. Since this scoring only depends on channel-level error and the fixed teacher reference, it can also be applied without modification to composite tasks despite the lack of composite task teachers. We therefore interpret the 0.5 baseline as the performance of an ideal model in which any task composition achieves per-channel tracking as good as the teachers.

Additionally, since all policies are trained with rewards that encourage foot-lifting patterns tied to their clock observations, we compute a gait error to quantify if a policy is performing a specified gait as intended. This is calculated as the error between expected foot contact patterns (derived from the sinusoidal clocks in the policy observations, see Appendix Section~\ref{subsec:gait_clock}) and actual contact patterns over a trial.

Finally, we compute a stance score. The stance score measures how many feet are in contact with the ground when $\norm{[\mathbf{v}^*, \boldsymbol{\omega}^*]} = 0$. By default, all feet on the ground earns a score of 1; 3 feet on the ground earns 0.5, and 2 or fewer feet on the ground earns 0. In the case when $\norm{\mathbf{L}^*} > 0$, 3 feet on the ground earns a score of 1.

{
\setlength{\tabcolsep}{5pt}
\begin{table*}[t]
\centering
\begin{threeparttable}
\caption{Comparison of Our Method Against DtF, PPO, and a Teacher Router}
\label{tab:main_results}
\renewcommand{\arraystretch}{1.2}
\begin{tabular}{l ccccccc ccccccc}
\toprule
& \multicolumn{7}{c}{Single Tasks} & \multicolumn{7}{c}{Composite Tasks}\\
\cmidrule(lr){2-8} \cmidrule(lr){9-15}
Method & SR [\%] & Exp $\mu$ & Imp $\mu$ & Exp $\sigma$ & Imp $\sigma$ & Gait Err & Stance & SR [\%] & Exp $\mu$ & Imp $\mu$ & Exp $\sigma$ & Imp $\sigma$ & Gait Err & Stance \\
\midrule
Teacher Router & 80.34 & 0.500 & 0.477 & 0.000 & 0.155 & 0.118 & 0.809 &  ---  &  --- &  ---  &  ---  &  ---  &  ---  &  ---  \\
\midrule
Ours          & 99.14 & \textbf{0.608} & 0.716 & \textbf{0.150} & 0.107 & \textbf{0.147} & \textbf{0.927} & \textbf{99.23} & \textbf{0.605} & 0.714 & \textbf{0.116} & 0.154 & \textbf{0.158} & \textbf{0.913} \\
DtF           & 96.99 & 0.474 & \textbf{0.741} & 0.249 & \textbf{0.092} & 0.383 & 0.895 & 96.96 & 0.496 & \textbf{0.731} & 0.132 & \textbf{0.132} & 0.384 & 0.867 \\
PPO           & \textbf{99.72} & 0.440 & 0.557 & 0.267 & 0.118 & 0.478 & 0.600 & \textbf{99.42} & 0.471 & 0.534 & 0.135 & 0.281 & 0.476 & 0.688 \\
\bottomrule
\end{tabular}
\begin{tablenotes}\footnotesize
\item SR: survival rate. Exp $\mu$, Imp $\mu$: explicit and implicit task score. Exp $\sigma$, Imp $\sigma$: explicit and implicit tracking standard deviation. The teacher router is omitted from composite tasks since it cannot perform them. Bold indicates the best value among the end-to-end policies; for SR, all values within 0.5\% of the best are bolded.
\end{tablenotes}
\end{threeparttable}
\vspace{-1em}
\end{table*}}
\subsection{Evaluation Procedure}

To measure policy performance, we conduct two tests across three seeds: one to evaluate task performance, motion quality, and robustness, and one to evaluate policy generalization.

In the first test, each model is commanded to perform a task $A \in \mathcal{Y}$ for 10 seconds and then another task $B \in \mathcal{Y}$ for 5 seconds. This is done for all possible combinations of $A$ and $B$, including cases where $A=B$. In such cases, we command a different tracking target. During evaluation, observation noise is turned on, and occasional external forces ranging from 0 to 300\,N are applied to the robot center-of-mass over a duration ranging from 0.1 to 0.2\,s. Each model receives the same command, disturbances, and observation noise. At each time step, we record the policy observation and evaluation channels, from which survival rate, explicit and implicit task scores, gait error, and stance score are computed. Scores are then averaged over all tasks within each segment, before the switch (A) and after the switch (B), to yield average performance scores. Task performance is extracted from segment $A$ and task switching performance is extracted from segment $B$. We run 40 trials per $A \to B$ sequence, yielding 19,360 total trials per model per seed.

In the generalization test, each model is given commands never observed during training, including 3-task and 4-task compositions and a novel 2-task composition. Observation noise and external force disturbance are configured identically to the first test. Metrics including survival rate, explicit task score, and implicit task score are tracked. In total, we test 7 out-of-distribution commands, with 60 trials each, yielding 420 trials per model per seed.

\section{Results and Discussions}

We aim to answer two primary research questions: (A) How does our policy perform on command tracking compared to baseline policies? (B) How much behavioral quality does our policy preserve from teachers compared to the baselines?

Overall, we find that our method performs comparably to or better than the baselines at all 22 tasks while exhibiting the highest motion quality and faithfulness with respect to the teachers. While some quantitative metrics show higher scores for baselines, we find that this is often due to the effects of reward gaming causing failure to actually perform the task where our method succeeds.

Additionally, we investigate these questions: (C) How well does our policy handle switching between tasks and tracking targets? (D) How well does our policy scale in terms of number of tasks learned? (E) How effective is adversarial task selection in improving policy performance? (F) How well does our policy generalize to OOD tasks? (G) How sensitive is our method to hyperparameter choice compared to the baselines? 

\begin{figure*}[t]
\centering
\includegraphics[width=17.5cm]{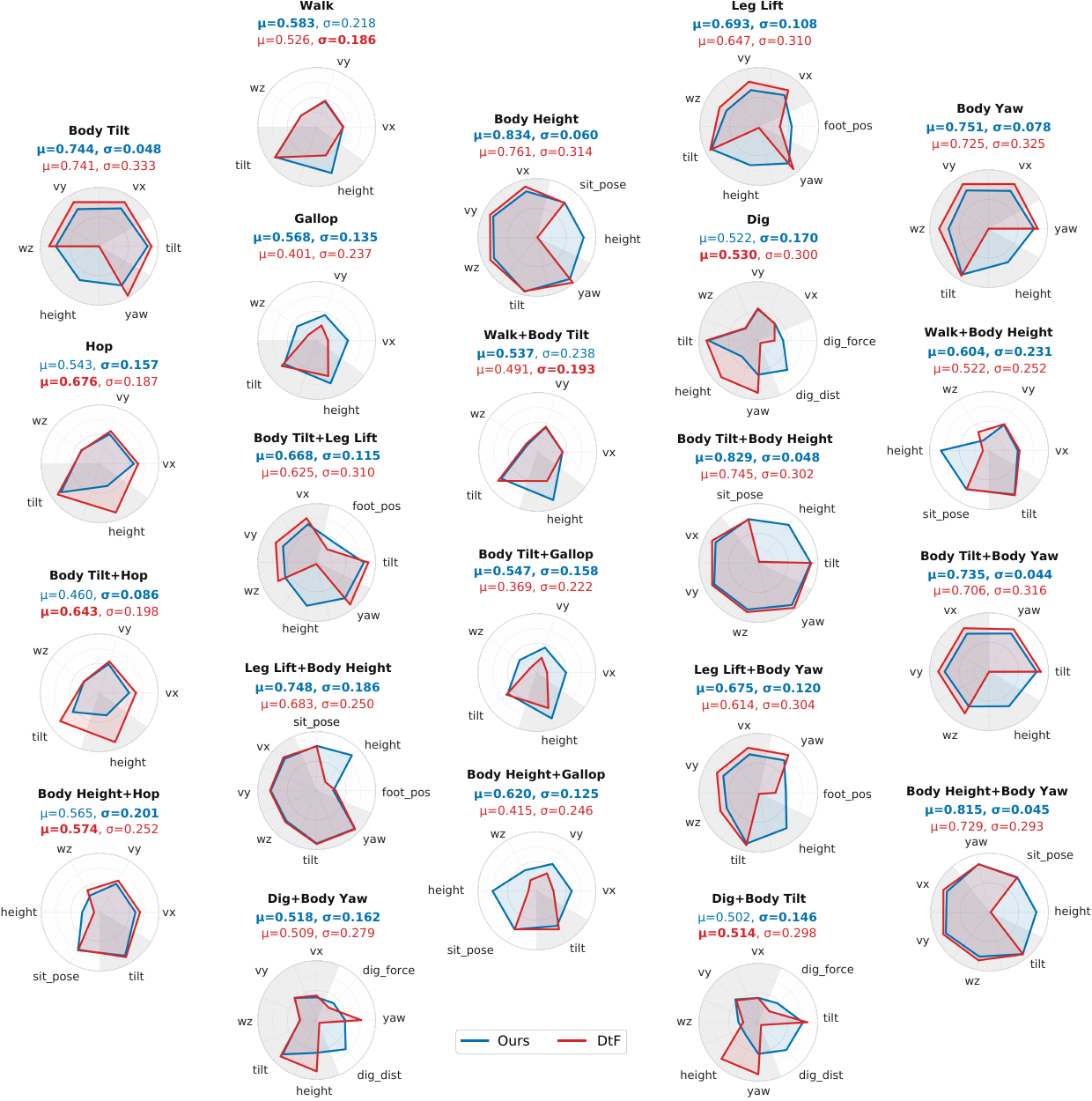}
\caption{Per-channel performance comparison of our method against DtF for each task. Top metrics (blue) are ours, bottom metrics (red) are DtF's. Each graph shows the performance breakdown for one task, and each axis shows the score for one evaluation channel (higher is better, outer ring is 1.0). Channels with a grey background are implicit. $\mu$ is the task's average score across its explicit and implicit channels, and $\sigma$ is the spread of scores across its channels.}
\label{fig:radar_task_channels}
\end{figure*}
\begin{figure*}[t]
\centering
\includegraphics[width=\textwidth]{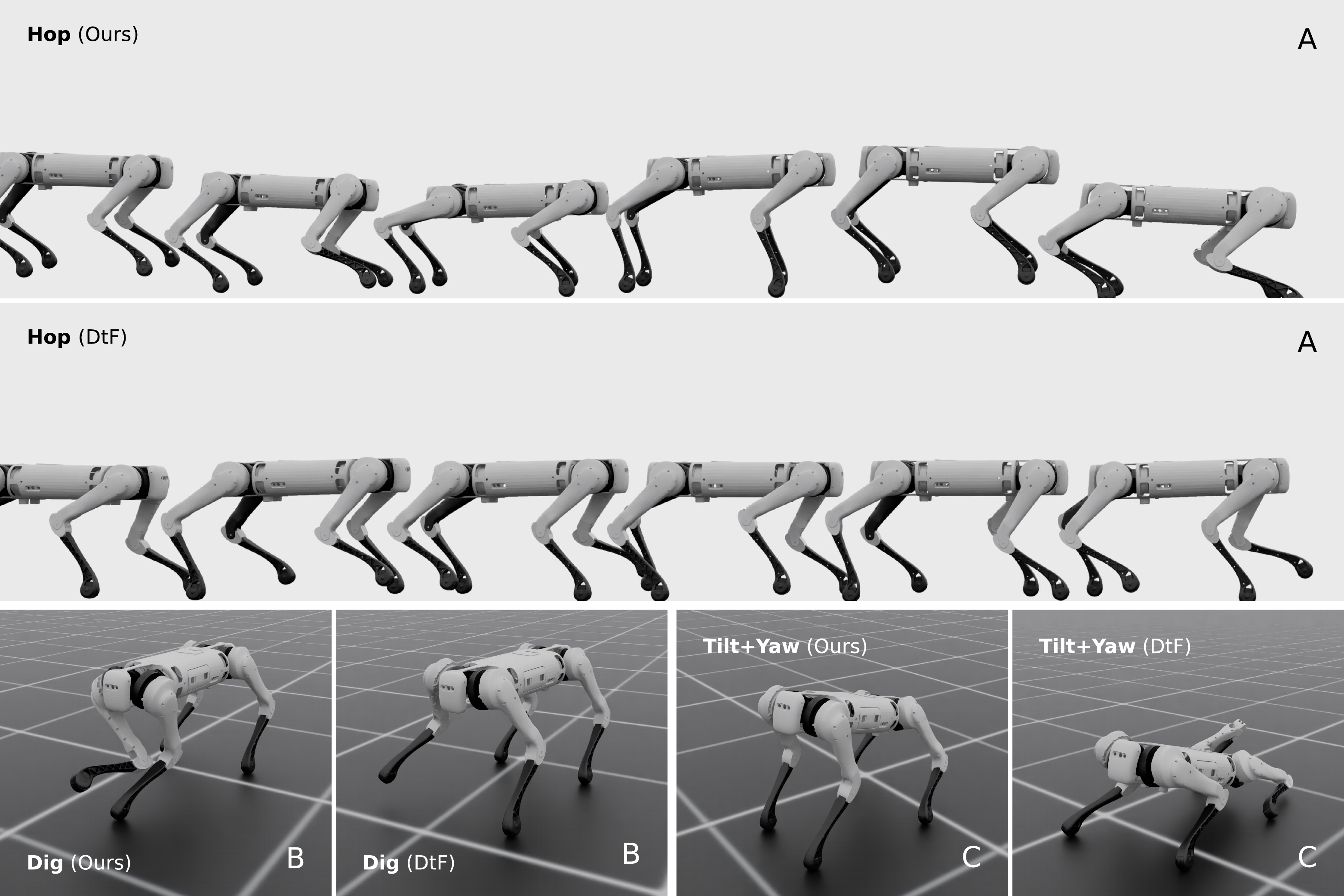}
\caption{Comparison of task execution between our method and DtF. (A) Our policy demonstrates a hop trajectory when commanded to hop, whereas DtF walks. (B) Our policy scrapes one shared area when commanded to dig, whereas DtF scrapes in two separate locations. (C) Our policy maintains the implicit height target during Tilt+Yaw, whereas DtF collapses to the ground.}
\label{fig:sim_robots}
\vspace{-1em}
\end{figure*}
\subsection{Multi-Task Learning}

When averaged across single and composite tasks, our policy exhibits a consistently higher explicit task score while maintaining the smallest explicit score spread, as shown in Table~\ref{tab:main_results}. It also achieves the smallest gait error and highest stance score, indicating movements that more closely match training intent. In contrast, a traditional PPO method proves ineffective, resulting in the lowest performance across almost all tasks.

Intriguingly, while our method outperforms DtF in explicit task scores, DtF obtains higher implicit task scores in both single and composite tasks. While this can suggest that DtF tracks held targets more stably, it can also indicate reward gaming: sacrificing performance on harder, changing targets to maximize reward on easier, constant ones. To test for this, we break down each task by per-channel performance for both methods in Fig.~\ref{fig:radar_task_channels}, where grey-background labels denote a task's implicit channels. Each task also shows the overall channel performance average and the STD of the channel scores. While our channel performance is relatively uniform across all 22 tasks, we see that DtF's is much more varied, shown by higher channel score STDs for all tasks except two. Often, DtF completely fails in certain channels while excelling in others --- such as in dig, where it sacrifices the explicit \texttt{dig\_force} and \texttt{dig\_dist} channels for dramatically improved scores on the implicit \texttt{height} and \texttt{yaw} channels. Implicit channels are also sacrificed --- such as in Tilt+Yaw, where the implicit \texttt{height} channel is sacrificed for improvements across all others. Indeed, as illustrated in Fig.~\ref{fig:sim_robots}, DtF fails to produce behaviors that it was trained for. When commanded to dig, DtF scrapes the front legs independently on each side instead of in one area. When commanded to Tilt+Yaw, DtF collapses to the ground despite the height target being held at 0.55 m. Most notably, while DtF appears to exhibit much stronger performance for hop and its composite tasks in Fig.~\ref{fig:radar_task_channels}, the movement trajectory of DtF in Fig.~\ref{fig:sim_robots} shows that it fails to produce a hopping motion when commanded to hop.

\begin{table}[t]
\centering
\begin{threeparttable}
\caption{Per-Task Teacher Action L2 Error and Aggregate Task Score on Single Tasks}
\label{tab:skill_tracking_vert}
\setlength{\aboverulesep}{0pt}
\setlength{\belowrulesep}{0pt}
\renewcommand{\arraystretch}{1.25}
\setlength{\tabcolsep}{8pt}  
\begin{tabular}{l ccc|ccc}
\toprule
& \multicolumn{3}{c|}{Teacher L2 Error} & \multicolumn{3}{c}{Task Score}\\
\cmidrule(lr){2-4} \cmidrule(lr){5-7}
Task    & Ours & DtF & PPO & Ours & DtF & PPO \\
\midrule
Walk    & \textbf{0.79} & 1.61 & 4.31 & \textbf{0.583} & 0.526 & 0.365 \\
Lift    & \textbf{0.76} & 3.57 & 3.99 & \textbf{0.693} & 0.647 & 0.562 \\
Tilt    & \textbf{0.64} & 2.94 & 3.83 & \textbf{0.744} & 0.741 & 0.637 \\
Height  & \textbf{0.73} & 1.45 & 2.63 & \textbf{0.834} & 0.761 & 0.661 \\
Yaw     & \textbf{0.75} & 3.26 & 2.67 & \textbf{0.751} & 0.725 & 0.626 \\
Gallop  & \textbf{0.94} & 2.70 & 5.80 & \textbf{0.568} & 0.401 & 0.269 \\
Dig     & \textbf{1.42} & 4.20 & 7.09 & 0.522 & 0.530 & \textbf{0.532} \\
Hop     & \textbf{0.85} & 2.25 & 4.93 & 0.543 & \textbf{0.676} & 0.435 \\
\bottomrule
\end{tabular}
\begin{tablenotes}\footnotesize
\item Lower error (left) indicates closer match to teacher behavior; higher score (right) indicates better overall task performance. Bold indicates the best value per task within each metric.
\end{tablenotes}
\end{threeparttable}
\vspace{-1em}
\end{table}
\subsection{Motion Quality}

Since the teacher policies are trained to produce motion we consider natural and desirable, we compare student actions to teacher actions to assess how well the student reproduces this desired motion quality. For each single task command, we sample the corresponding teacher policy under the same observation and command experienced by the student and compute the L2 error between the teacher and student actions. We also compute the overall task score in the same way as presented in Fig.~\ref{fig:radar_task_channels}. Each method's errors and scores are displayed in Table~\ref{tab:skill_tracking_vert}. Finally, we note that we do not consider metrics such as action rate or joint power, since a policy that fails to follow a commanded behavior (e.g., collapsing to the ground as demonstrated in Fig.~\ref{fig:sim_robots}) trivially attains lower values on these metrics.

When comparing teacher action errors, our method attains lower errors than either DtF or PPO, while DtF achieves better overall task scores on 2 of the 8 single tasks. However, as discussed previously, a higher task score does not necessarily indicate that the task is performed correctly. Indeed, on all tasks where DtF outscores our method, it does so at a much higher action error, deviating from motions we consider desirable. This is evident when comparing the action error and score of hop against the trajectories illustrated in Fig.~\ref{fig:sim_robots}, where DtF fails to produce a hopping motion altogether, instead opting for a walking gait that yields a higher overall score. Additionally, it fails to dig properly and holds a more stable stance with its feet apart. Together with the strong task tracking demonstrated previously, we show that our method can learn many tasks and track them well while preserving the high motion quality of the teacher policies.

\begin{table}[t]
\centering
\begin{threeparttable}
\caption{Task and Target Switching Performance}
\label{tab:task_switching}
\renewcommand{\arraystretch}{1.35}
\setlength{\aboverulesep}{0pt}
\setlength{\belowrulesep}{0pt}
\begin{tabular}{l ccc|ccc}
\toprule
& \multicolumn{3}{c|}{A$\rightarrow$A} & \multicolumn{3}{c}{A$\rightarrow$B}\\
\cmidrule(lr){2-4} \cmidrule(lr){5-7}
Model   & SR [\%] & Exp $\mu$ & Imp $\mu$ & SR [\%] & Exp $\mu$ & Imp $\mu$ \\
\midrule
Router  & 91.15 & 0.518 & 0.478 & 80.82 & 0.394 & 0.356 \\
\midrule
Ours    & \textbf{99.47} & \textbf{0.579} & 0.702 & 99.05 & \textbf{0.558} & 0.651 \\
DtF     & 97.74 & 0.466 & \textbf{0.724} & 97.66 & 0.395 & \textbf{0.695} \\
PPO     & \textbf{99.62} & 0.449 & 0.542 & \textbf{99.81} & 0.448 & 0.526 \\
\bottomrule
\end{tabular}
\begin{tablenotes}\footnotesize
\item SR: survival rate. Exp $\mu$, Imp $\mu$: explicit and implicit task score. Bold indicates survival rates within 0.5\% of the highest, and the highest explicit or implicit score.
\end{tablenotes}
\end{threeparttable}
\vspace{-1em}
\end{table}
\subsection{Task and Target Switching}
When switching between tasks or tracking targets, we expect an effective policy to respond promptly while remaining stable throughout the transition. To test this, we use segment B of the previously described testing procedure, reporting survival rates along with explicit and implicit task scores for both target switches (A$\rightarrow$A) and task switches (A$\rightarrow$B) given segment A is survived. We show these figures in Table~\ref{tab:task_switching}.

Our method achieves the highest explicit task score and maintains a survival rate above 99\% in both conditions. In contrast, the teacher router only attains a survival rate of 91.15\% when switching tracking targets and 80.82\% when switching tasks. This degradation likely stems from the router handing control between independently trained teachers at the switch, producing action discontinuities that destabilize the robot. PPO attains marginally higher survival rates. As established in Table~\ref{tab:main_results} and Table~\ref{tab:skill_tracking_vert}, it neither tracks targets nor produces desirable motion, indicating that this stability comes from ignoring commands rather than executing them. DtF earns the highest implicit score, but as discussed previously, this does not necessarily reflect better task performance.

\begin{table}[t]
\centering
\begin{threeparttable}
\caption{Evaluation of Task Scaling}
\label{tab:task_scaling}
\renewcommand{\arraystretch}{1.25}
\setlength{\aboverulesep}{0pt}
\setlength{\belowrulesep}{0pt}
\setlength{\tabcolsep}{3.0pt}  
\begin{tabularx}{\columnwidth}{>{\raggedright\arraybackslash}X cccc | cc}
\toprule
& \multicolumn{4}{c|}{9 Tasks} & \multicolumn{2}{c}{22 Tasks}\\
\cmidrule(lr){2-5} \cmidrule(lr){6-7}
Policy & Ours (9) & DtF (9) & Ours (22) & DtF (22) & Ours (22) & DtF (22) \\
\midrule
\textit{Single} & & & & & & \\
Exp $\mu$ & \textit{0.694} & \textit{0.679} & \textbf{0.650} & 0.523 & \textbf{0.608} & 0.474 \\
Imp $\mu$ & \textit{0.798} & \textit{0.812} & \textbf{0.794} & 0.754 & 0.716 & \textbf{0.741} \\
Gait Err  & \textit{0.159} & \textit{0.139} & 0.178 & \textbf{0.154} & \textbf{0.147} & 0.383 \\
Stance    & \textit{1.000} & \textit{0.999} & \textbf{1.000} & \textbf{1.000} & \textbf{0.927} & 0.895 \\
\midrule
\textit{Composite} & & & & & & \\
Exp $\mu$ & \textit{0.676} & \textit{0.651} & \textbf{0.627} & 0.513 & \textbf{0.605} & 0.496 \\
Imp $\mu$ & \textit{0.817} & \textit{0.811} & \textbf{0.819} & 0.762 & 0.714 & \textbf{0.731} \\
Gait Err  & \textit{0.160} & \textit{0.149} & 0.179 & \textbf{0.164} & \textbf{0.158} & 0.384 \\
Stance    & \textit{1.000} & \textit{0.999} & \textbf{1.000} & \textbf{1.000} & \textbf{0.913} & 0.867 \\
\bottomrule
\end{tabularx}
\begin{tablenotes}\footnotesize
\item Exp $\mu$, Imp $\mu$: explicit and implicit task score. 9-task models are evaluated on the 9-task subset; 22-task models are evaluated on both the 9-task subset and the full 22 tasks. Parenthesis next to the policy name indicates the number of tasks it was trained to perform. Bold indicates no or smaller 9-to-22 degradation between ours and DtF.

\end{tablenotes}
\end{threeparttable}
\vspace{-1em}
\end{table}
\subsection{Task Scaling}
To assess how well our method scales to a growing number of tasks, we train two additional models, one using DtF and one using our method, on a reduced set of 9 tasks composed of 4 single tasks (Walk, Body Height, Body Tilt, and Leg Lift) and 5 composite tasks (Walk+Tilt, Walk+Height, Lift+Tilt, Tilt+Height, and Lift+Height). We evaluate the 9-task models on these 9 tasks and the original 22-task models on both the 9-task subset and the set of all 22 tasks. For each model and evaluation set, we report the overall task performance, gait error, and stance score across the evaluated tasks in Table~\ref{tab:task_scaling}. When trained to perform 9 tasks, we observe that DtF is competitive with our method under the 9-task subset, and even attains a slightly lower gait error. When trained to perform 22 tasks, however, DtF exhibits a much higher degradation in explicit and implicit scores under the 9-task subset, and suffers more degradation in all metrics except implicit scores under the 22-task set. While both methods degrade in performance on most metrics, DtF's higher degradation in explicit task scores under both the 9-task subset and 22-task set reflects a more significant loss in task performance as well as ineffectiveness in learning the new tasks. In contrast, our method suffers less degradation on most metrics across tasks when scaling from 9 to 22 tasks, indicating that our method learns more effectively than DtF when more tasks are added to training.

\begin{figure}[t!]
\centering
\includegraphics[width=7.5cm]{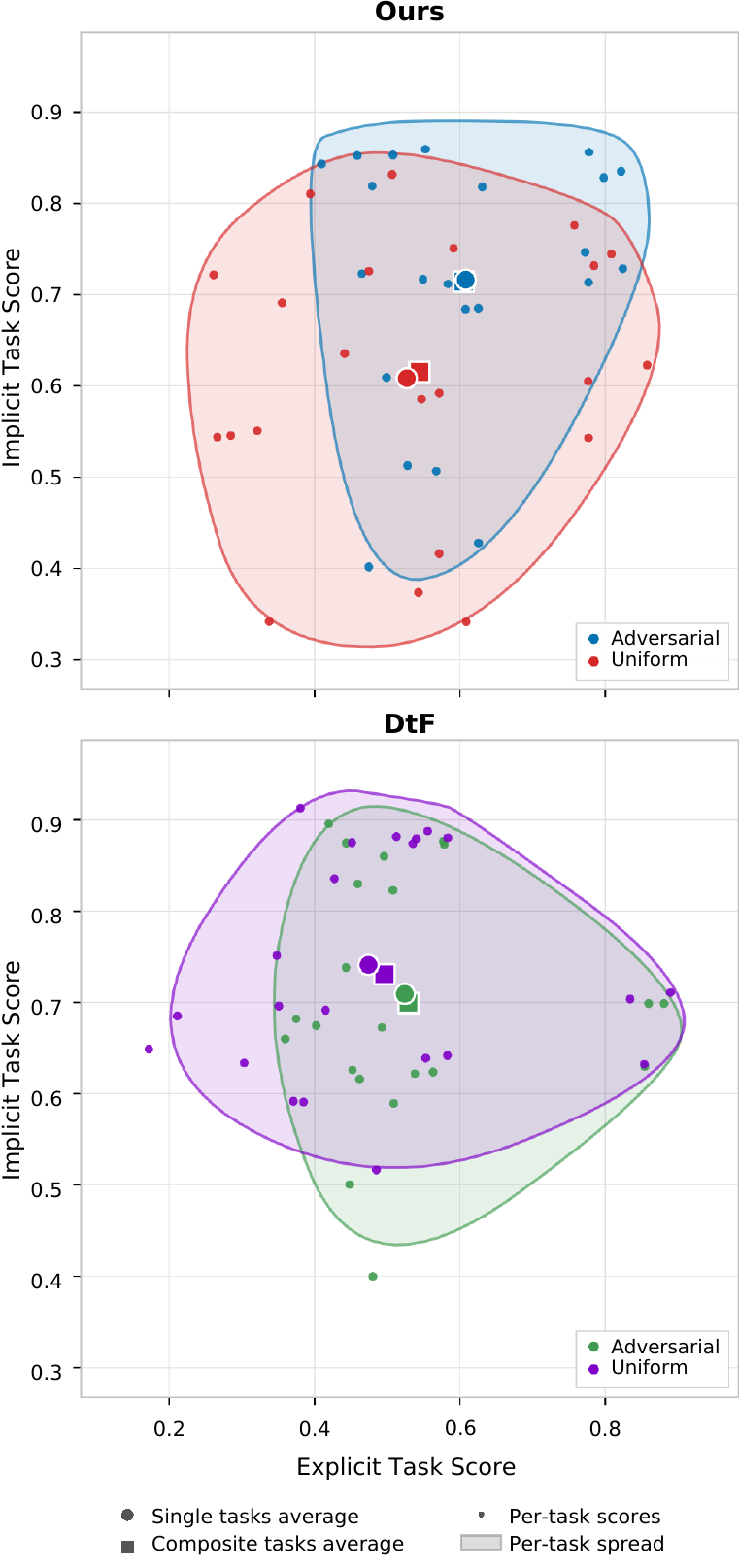}
\caption{Visualizations of adversarial task selection's effect on implicit and explicit task scores across single and composite tasks for our method and DtF. Individual task scores are also shown, with their spread shaded.}
\label{fig:adversarial_scatter}
\vspace{-1em}
\end{figure}
\subsection{Adversarial Task Selection}

To assess the effectiveness of adversarial task selection, we train two additional models: one using our method without adversarial task selection, and one using DtF with adversarial task selection. The resulting task scores are shown in Fig.~\ref{fig:adversarial_scatter}. For both methods, adversarial task selection improves the mean explicit task score for both single and composite tasks while tightening the score spread across tasks. This behavior follows the intention of the adversarial mechanism, where tasks that perform poorly under uniform sampling are targeted more frequently and thus improve more, while tasks that already perform well degrade only marginally. Since adversarial task selection evaluates task performance through error channels similar to those used in our explicit evaluation, we see improvements in explicit task scores for both methods. Notably, while adversarial task selection also improves implicit task scores for our method, it degrades them for DtF. This effect suggests that without the BC term in our optimization objective, adversarial task selection can only trade performance between explicit and implicit channels.

\begin{table*}[t]
\centering
\begin{threeparttable}
\caption{Ablation of Initial Policy Standard Deviation and Learning Rate}
\label{tab:hyperparameter_ablation}
\renewcommand{\arraystretch}{1.25}
\setlength{\tabcolsep}{5.0pt}  
\begin{tabular}{l ccccccc ccccccc}
\toprule
& \multicolumn{7}{c}{Single Tasks} & \multicolumn{7}{c}{Composite Tasks}\\
\cmidrule(lr){2-8} \cmidrule(lr){9-15}
Method & SR [\%] & Exp $\mu$ & Imp $\mu$ & Exp $\sigma$ & Imp $\sigma$ & Gait Err & Stance & SR [\%] & Exp $\mu$ & Imp $\mu$ & Exp $\sigma$ & Imp $\sigma$ & Gait Err & Stance \\
\midrule
Ours 1.0 STD  & \textbf{99.80} & \textbf{0.595} & \textbf{0.698} & \textbf{0.172} & \textbf{0.105} & \textbf{0.148} & 0.922 & \textbf{99.90} & \textbf{0.600} & \textbf{0.704} & \textbf{0.127} & \textbf{0.139} & \textbf{0.164} & 0.905 \\
DtF 1.0 STD   & \textbf{99.99} & 0.380 & 0.648 & 0.267 & 0.157 & 0.432 & \textbf{1.000} & \textbf{99.99} & 0.418 & 0.638 & 0.142 & 0.278 & 0.427 & \textbf{1.000} \\
\midrule
Ours LR1e-4    & \textbf{99.72} & \textbf{0.537} & 0.632 & \textbf{0.216} & 0.103 & \textbf{0.189} & 0.706 & \textbf{99.44} & \textbf{0.550} & 0.636 & 0.156 & 0.150 & 0.255 & 0.724 \\
DtF  LR1e-4    & 97.62 & 0.511 & \textbf{0.745} & 0.245 & \textbf{0.085} & 0.382 & \textbf{0.907} & 97.28 & 0.535 & \textbf{0.736} & 0.131 & \textbf{0.128} & 0.380 & \textbf{0.877} \\
Ours LR2e-5    & \textbf{99.72} & 0.525 & 0.615 & 0.231 & 0.103 & 0.193 & 0.697 & \textbf{99.73} & 0.542 & 0.602 & 0.161 & 0.150 & \textbf{0.242} & 0.729 \\
DtF  LR2e-5    & 98.96 & 0.443 & 0.719 & 0.247 & 0.104 & 0.377 & 0.866 & 98.87 & 0.466 & 0.692 & \textbf{0.127} & 0.162 & 0.380 & 0.840 \\
\bottomrule
\end{tabular}
\begin{tablenotes}\footnotesize
\item Metrics are as defined in Table~\ref{tab:main_results}. STD: initial policy standard deviation. LR: initial learning rate. Bold indicates the best value per metric within each group; for SR, all values within 0.5\% of the best are bolded.
\end{tablenotes}
\end{threeparttable}
\vspace{-1em}
\end{table*}
\begin{table}[t]
\centering
\begin{threeparttable}
\caption{Evaluation on Out-of-Distribution Commands}
\label{tab:task_generalization}
\renewcommand{\arraystretch}{1.25}
\setlength{\tabcolsep}{3pt}  
\begin{tabular}{l ccc ccc}
\toprule
& \multicolumn{3}{c}{Ours} & \multicolumn{3}{c}{DtF}\\
\cmidrule(lr){2-4} \cmidrule(lr){5-7}
Task & SR [\%] & Exp $\mu$ & Imp $\mu$ & SR [\%] & Exp $\mu$ & Imp $\mu$ \\
\midrule
Dig+Height           & 96.35 & \textbf{0.441} & 0.496 & \textbf{100.00} & 0.291 & \textbf{0.617}\\
Dig+Tilt+Yaw         & 98.81 & \textbf{0.556} & 0.418 & \textbf{100.00} & 0.459 & \textbf{0.494}\\
Lift+Tilt+Yaw        & 98.57 & 0.592 & \textbf{0.673} & \textbf{99.84} & \textbf{0.618} & 0.566\\
Tilt+Height+Gallop   & \textbf{99.44} & \textbf{0.580} & --- & 83.02 & 0.398 & --- \\
Tilt+Height+Yaw      & \textbf{99.92} & \textbf{0.802} & 0.803 & \textbf{100.00} & 0.616 & \textbf{0.844}\\
Walk+Tilt+Height     & \textbf{100.00} & \textbf{0.546} & --- & \textbf{100.00} & 0.483 & --- \\
Lift+Tilt+Height+Yaw & \textbf{99.44} & \textbf{0.665} & 0.738 & \textbf{99.92} & 0.554 & \textbf{0.763}\\
\bottomrule
\end{tabular}
\begin{tablenotes}\footnotesize
\item SR: survival rate. Exp $\mu$, Imp $\mu$: explicit and implicit task score. Implicit score is omitted for compositions without implicit targets. Bold indicates survival rates within 0.5\% of the highest and the highest task score.
\end{tablenotes}
\end{threeparttable}
\vspace{-1em}
\end{table}
\subsection{Generalization}
An important aspect of policy learning is the ability to generalize to OOD inputs. To evaluate the generalization of each method, we supply commands never seen during training. These include the composite task Dig+Height as well as a series of more complex composite tasks composed of 3 or 4 single tasks. The tasks and their performance are reported in Table~\ref{tab:task_generalization}. For 6 of the 7 tasks, our method attains an explicit task score that is higher than DtF's, whereas DtF survives more often and attains a higher implicit task score overall. As discussed in previous sections, however, higher survival rate and implicit score do not necessarily indicate good task performance. This is demonstrated clearly on Tilt+Height+Gallop, which has no implicit channels to measure. DtF performs below teacher baselines while our method remains above it. DtF's survival rate on this task is also lower than ours, suggesting a failure to generalize dynamic motion, whereas our method achieves a near-perfect survival rate.

\subsection{Hyperparameter Choices}
During training, we also find that our method exhibits greater robustness to hyperparameter choices than DtF in certain aspects. Most notably, while a high initial STD during finetuning causes significant performance degradation for DtF, as shown in Table~\ref{tab:hyperparameter_ablation}, it has minimal effects on our method when applied to the RLP stage. We attribute this to the BC term, which always supplies a gradient toward predictable teacher actions and thus acts as a safe ``anchor'' during chaotic exploration. As DtF has no such gradient, aggressive exploration destabilizes the policy and degrades its performance. Since a higher initial STD encourages exploration, we speculate that this anchoring gradient may also let our method discover more novel or stable solutions.

We also ablate the initial learning rate in Table~\ref{tab:hyperparameter_ablation}. Our method degrades under both alternative choices, whereas DtF improves its explicit score at a larger initial LR, suggesting that DtF may require a higher LR to learn explicit command tracking. However, its gait error remains large and unchanged, indicating that the gait failures shown in Fig.~\ref{fig:sim_robots} persist. The overall performance trends across methods are unchanged, and the higher initial LR did not meaningfully improve DtF's gait error or stance score on either single or composite tasks.

From these ablations, we conclude that while our method is more resilient to higher initial STD values, it is also more sensitive to the choice of initial learning rate.

\section{Real-World Deployment}

To achieve robust real-world deployment, we apply standard domain randomization techniques during training, including observation noise and delays. Each observation channel's noise is modeled as a Gaussian with a bias and variance, where the bias is sampled uniformly from a fixed range at the start of each episode. We additionally model external force disturbances as forces applied to the robot's center of mass over a randomized time duration. We deploy the resulting policy on a standard Unitree B1 quadruped using only proprioceptive observations; no depth cameras or other exteroceptive sensors are used. Policy inference happens onboard on a Raspberry Pi 4B at 100\,Hz. We evaluate the robot indoors and outdoors on flat concrete, grass fields, and loose dirt, where it successfully crawls under tight spaces, opens doors, pushes a battery into a dead robot, and digs up the ground. These results are shown in Fig.~\ref{fig:irl_robots} and the video at \textbf{\href{https://youtu.be/V9yX04EBcFA}{https://youtu.be/V9yX04EBcFA}}.

\section{Conclusion}

In this work, we introduce a three-stage MTRL training method designed to address the challenges of effective and diverse task learning with RL. By optimizing a combined IL and RL objective and sampling tasks under adversarial task selection, we enable effective multi-task learning within a single end-to-end policy. Experimental results demonstrate that policies trained under this method track command targets without sacrificing motion quality or resorting to excessive reward gaming, while exhibiting less performance degradation as the number of learned tasks grows and being able to generalize to tasks not seen during training. 

Since this work relies solely on proprioceptive observations, a natural next step is to incorporate exteroceptive sensing, extending the learned tasks and their composition to terrain- and object-aware behaviors. We also leave evaluation of the method's effectiveness on other robot embodiments and policy architectures, such as recurrent models, to future work.
{\appendix

\subsection{Training Setup}
\label{subsec:training_setup}
Each teacher model is a three-layer MLP containing 256, 128, and 64 neurons per layer. The student policy (under our method, DtF, or PPO) uses a similarly structured three-layer MLP with 512, 256, and 128 neurons per layer. For adversarial task selection, we set its hyperparameters at constant values of $\kappa = 0.85$ and $\beta = 1.5$ throughout training. During ILP, we use $\lambda_\mathrm{IL} = 1.0$, and during RLP, we use $\lambda_\mathrm{RL}=0.25$. 

All models are trained in a flat-terrain environment with identical domain randomization in NVIDIA Isaac Lab \cite{nvidia2025isaaclabgpuacceleratedsimulation}. Each method's RL segment is trained for 1.229 billion transitions under an identical reward landscape; for PPO and the finetuning stage of DtF, the RLP reward landscape is used. By default, we use an initial STD of 0.5 during the RLP stage of our method and the finetuning stage of DtF, and 1.0 for traditional PPO. All methods use a default initial learning rate of $5 \times 10^{-5}$, decaying linearly to $5 \times 10^{-6}$ over the first $80\%$ of training. During ILP, tasks are held constant for the duration of each episode while tracking targets are varied. During RLP, tasks and tracking targets switch mid-episode.

\subsection{Gait Clock}
\label{subsec:gait_clock}
To encourage more consistent gait structures, we add two sinusoidal phase clocks to all proprioceptive policy observations. Each clock is a $(\sin, \cos)$ pair at frequencies $f_1 = 1.7\,\text{Hz}$ and $f_2 = 3.4\,\text{Hz}$, yielding four additional observation terms
\begin{equation*}
[\sin(2\pi f_1 t),\, \cos(2\pi f_1 t),\, \sin(2\pi f_2 t),\, \cos(2\pi f_2 t)],
\end{equation*}
where $t$ is the elapsed episode time. Teacher policies may also receive a single clock pair at $f_1$ or $f_2$ when the task requires a gait such as walk. In our tasks, walk and hop receive $f_1$ and gallop receives $f_2$, locking each of these gaits to a predictable footfall pattern. The student observes both clocks.

}

\bibliographystyle{IEEEtran}
\bibliography{references} 

\end{document}